\documentclass[journal]{IEEEtran}

\ifCLASSINFOpdf

\else

\fi

\usepackage{amsmath}
\usepackage{amsfonts}
\usepackage{amssymb}
\usepackage{graphicx}
\usepackage[left=2cm,right=2cm,top=2cm,bottom=2cm]{geometry}
\usepackage{subcaption}
\usepackage{caption}
\usepackage{cite}
\usepackage{nth}
\usepackage{flexisym}
\usepackage[export]{adjustbox}

\usepackage{mathtools}
\usepackage{dsfont}
\usepackage{multirow}
\usepackage{graphicx}
\usepackage[table]{xcolor}
\usepackage{algorithm}
\usepackage{algorithmic}

\usepackage{color}

\usepackage{multirow}

\newcommand{\Sec}[1]{\hyperref[sec:#1]{\S\ref*{sec:#1}}} %section
\newcommand{\Eqn}[1]{\hyperref[eq:#1]{(\ref*{eq:#1})}} %equation
\newcommand{\Fig}[1]{\hyperref[fig:#1]{Figure~\ref*{fig:#1}}} %figure
\newcommand{\Tab}[1]{\hyperref[tab:#1]{Table~\ref*{tab:#1}}} %table
\newcommand{\Thm}[1]{\hyperref[thm:#1]{Theorem~\ref*{thm:#1}}} %theorem
\newcommand{\Lem}[1]{\hyperref[lem:#1]{Lemma~\ref*{lem:#1}}} %lemma
\newcommand{\Prop}[1]{\hyperref[prop:#1]{Property~\ref*{prop:#1}}} %property
\newcommand{\Cor}[1]{\hyperref[cor:#1]{Corollary~\ref*{cor:#1}}} %corollary
\newcommand{\Def}[1]{\hyperref[def:#1]{Definition~\ref*{def:#1}}} %definition
\newcommand{\Alg}[1]{\hyperref[alg:#1]{Algorithm~\ref*{alg:#1}}} %algorithm
\newcommand{\Ex}[1]{\hyperref[ex:#1]{Example~\ref*{ex:#1}}} %example

\newcommand{\mb}[1]{\mathbb{#1}}

\newcommand{\V}[1]{{\bm{\mathbf{\MakeLowercase{#1}}}}} % vector
\newcommand{\M}[1]{{\bm{\mathbf{\MakeUppercase{#1}}}}} % matrix
\newcommand{\T}[1]{\boldsymbol{\mathscr{\MakeUppercase{#1}}}} %tensor
\usepackage[mathscr]{eucal}

\usepackage{amsbsy}

\usepackage{bm}
\newtheorem{theorem}{Theorem}[section]

\newtheorem{defn}[theorem]{Definition}

\usepackage{amsfonts}

\def\HMSL{Hierarchical Subspace Learning }

\begin{document}
\title{Sample, Computation vs Storage Tradeoffs for Classification Using Tensor Subspace Models \thanks{{The authors are with the Dept. of ECE, Tufts University, Medford, MA. They can be reached at mohammad.chaghazardi@tufts.edu and shuchin@ece.tufts.edu. This research is supported by NSF:CCF:1319653 and NSF:CCF:1553075}}}
\author{
\IEEEauthorblockN{Mohammadhossein Chaghazardi and Shuchin Aeron, \emph{Senior Member, IEEE}}
%\IEEEauthorblockA{Dept. of Electrical and Computer Engineering, Tufts University, Medford, MA}
}

\maketitle

\begin{abstract}
In this paper\textcolor{black}{,} we exhibit the tradeoffs between (training) sample, computation and storage complexity for the problem of supervised classification using signal subspace estimation. Our main tool is the use of tensor subspaces, i.e. subspaces with a Kronecker structure, for embedding the data into lower dimensions. Among the subspaces with a Kronecker structure, we show that using subspaces with a hierarchical structure for representing data leads to improved tradeoffs. One of the main reasons for the improvement is that embedding data into these hierarchical Kronecker structured subspaces prevents overfitting at higher latent dimensions.
\end{abstract}

\vspace{-1.0em}
\section{Introduction}
The principle of dimensionality reduction is important for many machine learning and statistical signal processing tasks. The simplest of these approaches, embeds the data into a low-dimensional linear subspace or a locally linear subspace. In this paper we exploit a sub-class of subspaces that have a tensor or Kronecker structure, namely that they are constructed out of tensor product of other low-dimensional subspaces. Among these we study the Tucker subspace \cite{tuckerals-1980}, the Hierarchical-Tucker (HT) subspace \cite{hk-ht-2009}, and the Tensor-Train (TT) \cite{Oseledets_SIAM2011} subspace models, though further generalizations are possible\footnote{TT subspaces are a special case of HT subspaces. But for the sake of clarity we will treat them as two different cases in this paper.}. It is to be noted that finding the optimal tensor subspace representation in general is a computationally hard problem, although there exist several efficient algorithms \cite{Lathauwer_SIAM2000}. In this paper we consider the approximations obtained by variations of the higher order and hierarchical singular value decomposition algorithms for finding tensor subspace representations \cite{Lathauwer_SIAM2000,gras-hsvd-2010}. The main objective of the paper is to numerically study the tradeoffs between the sample complexity, computational cost of projection, storage (of the subspace representation) and error tradeoffs when using tensor subspaces for supervised classification \cite{Vasilescu2003,LU20111540,Sidiropoulos,DBLP:journals/corr/CichockiMPCZZL14}. While implicit, to the best of our knowledge, these tradeoffs are brought to attention for the first time via a direct numerical study.

\textbf{Main results}: For a fixed classification error, in this paper we note the following points regarding the tradeoffs. 
\begin{itemize}
\item The storage complexity of the HT subspace is much higher compared to the Tucker subspace. This is not surprising since specification of HT subspace requires more parameters \footnote{Technically speaking one can further reduce the cost of storage of the HT Subspace representations by considering the overall dimension of the subspace obtained by quotienting out equivalent representations \cite{USCHMAJEW2013133}. However this does not reduce the orders and using this optimally compressed representation dramatically increases the projection costs by requiring a representation to be computed explicitly.}. On the other hand the total cost of the projection onto the tensor subspace is lower for the HT compared to the Tucker model. 
\item The sample complexity, i.e. the number of data examples required to learn the subspace, is lower for HT compared to the Tucker. This is due to the fact that the overall dimension of the manifold of fixed rank HT subspace (polynomial in tensor order) is much smaller compared to the overall dimension of the manifold of fixed rank Tucker subspace (exponential in tensor order).    
\end{itemize}

The rest of the paper is organized as follows. In section \ref{sec:2} we provide necessary notation and technical background. In section \ref{sec:3} we outline a simple algorithm for finding a hierarchical Tucker subspace fitting the data. In section \ref{sec:4}, we show numerical results for the problem of supervised classification and highlight the variety of tradeoffs between classification error, storage and computational complexity that can be achieved using these subspaces on a variety of image databases. 
%There are several theoretically interesting questions that one can pose based on these observations. 

 %Then we proceed to modeling the problem of solving hierarchical structures. We propose our algorithm for solving the problem. Finally, in the results sections we compare different algorithms on real-world data sets and investigate the results.

% In addition, the algorithms of subspace classification often do not perform well and this triggers the point whether current algorithms are capturing the correct structure of the data. The idea is subspace bears the property of hierarchical nestedness; the subspace itself is a subspace of several other lower dimensional subspaces\cite{HS_thesis}. The algorithm poposed in this paper, solves the inner structure of a \textit{Hierarchical Subspace} which could represent the class of data points. 

\section{Notation and background}
\label{sec:2}

Through out this paper, we denote the set of integers $\{1,\cdots,N \}$ via $[N]$. 
We denote the size of the set $S$ via $|S|$.
We denote vectors with small bold-face letters like $\V{b}$, matrices by capital bold-face letters like $\M{B}$. The $i^{th}$ column of a matrix $\M{b}$ is denoted by $(\M{b})_i$ and the $(i,j)$ element is denoted by $\M{b}_{ij}$. Matrix fibers are extracted by using the colon notation. For example, columns of $i$ to $j$ are denoted by $\M{B}(:, i:j)$. We depict trees with $\mathfrak{T}$ symbols. 
Tensors are denoted by bold-face calligraphic letters like $\T{X} \in \mathbb{R}^{I_1 \times \cdots \times I_n}$, where $I_i$ is the size along the $i^{th}$ \emph{dimension/direction} of the tensor and $n$ denotes the order of the tensor. 

\textbf{Tensor Subspaces}: Most of the development has been distilled from \cite{Hackbusch_book}. All subspaces in this paper are subspaces of $\mb{R}^{d}$ of appropriate dimension $d$. The dimension $d$ will be clear from the context. Recall that a single subspace $\mathbb{S}$ of dimension $r$ can be expressed as, 
$$ \mathbb{S}= col-span\{\M{U}\} $$
for some rank-$r$ matrix $\M{U}$. Given two subspaces $\mathbb{S}_1 = col-span\{\M{U}_1 \in \mathbb{R}^{I_1 \times r_{1}} \} ,  \mathbb{S}_2 = col-span\{\M{U}_2  \in \mathbb{R}^{I_2 \times r_{2}}\}$, a tensor subspace denoted $\mathbb{S}_1 \otimes \mathbb{S}_2$ is defined via, %
%
%A single-level \textcolor{red}{Hierarchical Subspace} $\mb{HS}$ is defined via tensor product of two subspaces $\mb{S}_1,\mb{S}_2$ of dimensions $r_1,r_2$, like so,
\begin{equation*}
%  \begin{aligned}
%  \mathbb{S}_1 &= col-span\{\M{U}_1 \in \mathbb{R}^{I_1 \times r_{1}} \}\\
%  \mathbb{S}_2 &= col-span\{\M{U}_2  \in \mathbb{R}^{I_2 \times r_{2}}\}\\
   \mathbb{S}_1 \otimes \mathbb{S}_2 = col-span(\M{U_1} \otimes \M{U_2}),
    %\end{aligned}
\end{equation*}    
%where
%\begin{equation}    
% \quad (\M{U_{1,2}})_i = \sum_{j=1}^{r_1}\sum_{l=1}^{r_2} ((\M{U_1})_j \otimes (\M{U_2})_l)\T{B}_{jli}
%  \label{eq:SingleHierarchy}
%\end{equation}

%Here $\T{B}$ is referred to as the \textit{transfer tensor}, which plays the role of choosing the subspace of the tensor subspace $\mb{S}_1\otimes \mb{S}_2 = col-span(\M{U_1} \otimes \M{U_2})$. Equation (\ref{eq:SingleHierarchy}) can also be written as matrix multiplication,
%\begin{equation}
%\M{u}_{1,2}=(\M{U}_1 \otimes \M{U}_2)\M{b}
%\label{eq:matrixHierarhy}
%\end{equation}
%where $\M{b} \in \mathbb{R}^{I_1I_2 \times r_{1,2}}$ is a full-rank matrix with rank $r_{1,2}$. This implies that the rank of the subspace $\mb{HS}$ is $r_{1,2}$. Note that $\M{U}_{1,2}$ is not necessarily an orthonormal matrix.  As seen in the equation \eqref{eq:matrixHierarhy}, a hierarchical subspace is defined by both hidden subspaces and the structure of the transfer tensor. 
This \emph{construction} can also be naturally extended to a collection of subspaces $\mb{S}_1,...,\mb{S}_n$ yielding $ \bigotimes_{i=1}^{n} \mb{S}_i = col-span(\bigotimes_{i=1}^{n} \M{U}_i)$, where the notation $\bigotimes_{i=1}^{n}$ is a short-hand for tensor/Kronecker products. % For a matrix $\M{U}_{1,2} \in \mathbb{HS}$, we can observe how it can be generated from low-level matrices of $\mathbb{S}_1$ and $\mathbb{S}_2$. From equation (\ref{eq:matrixHierarhy}) one can write:
%$$\text{rank}(\M{U}_{1,2}) = \text{rank}(\M{B})=r_{1,2} $$
%\begin{defn} Given base subspaces $\mb{S}_i = \M{U}_i$, $i\in \{1,\cdots, n\}$ with ranks $r_i$ and a transfer tensor, $\T{B} \in \mathbb{R}^{r_1 \times r_2 \times \cdots \times r_n \times r_{1,2,\cdots,n}}$, a one single level subspace is defined as the column span of the matrix $\M{U}_{1,2,\cdots,n}$ with columns given by
%\end{defn}
%\begin{align}
%\label{eq:TuckerDecompSum}
%&(\M{U}_{1,2,\cdots,n})_{i} \\
%&= \sum_{j_1=1}^{r_1} \sum_{j_2=1}^{r_2} \cdots \sum_{j_n=1}^{r_n}( (\M{U}_1)_{j_1}\otimes \cdots \otimes (\M{U}_n)_{j_n})  \T{B}_{j_1j_2j_3\cdots j_ni} \,\,\, .
%\end{align}
%One can think of transforming the transfer tensor  $\T{B}$ into a full rank matrix $\M{B}$ of size $r_1 r_2 \cdots r_n \times r_{1,2,...,n}$ and express
%\begin{align}
%\M{U}_{1,2,...,n} = (\M{U}_1\otimes \cdots \otimes \M{U}_n) \M{B}
%\end{align} 
Note that a single element say $\V{x}$ from this tensor product of subspaces can be expressed as,
%when expressed as a linear combination of the columns of $\M{U}_{1,2,..,n}$ can be expressed as 
\begin{align}
\V{X}_i = (\M{U}_1\otimes \cdots \otimes \M{U}_n) \V{c}
\end{align} 
where $\V{b}$ is a vector of size $r_1r_2\cdots r_n$. One can \emph{reshape} this vector into a core tensor $\T{C}$ of size $r_1\times \cdots \times r_n$, thereby obtaining what is referred to as the Tucker decomposition \cite{KoldaMatrix} of $\V{x}$ or of a tensor $\T{X}$ obtained by \emph{reshaping} $\V{x}$\footnote{This reshaping is a standard operation considered in all tensor literature and is the opposite of the vectorization operation.}. %\textcolor{black}{We note that $\V{c} = \M{B} \V{d}$ for some vector $\V{d}$}. %(\textcolor{red}{it might better to write  $\M{B} d_i$ where $d_i$ is an arbitrary vector. Core tensor was called \T{B} in above equations} see \ref{eq:HiCore}). 

\textbf{Hierarchical tensor subspaces} - Instead of a single level construction, one can construct tensor subspaces in a hierarchical manner. These subspaces are referred to as Hierarchical Tensor Subspaces  or Hierarchical Tucker (HT) subspaces  \cite{gras-hsvd-2010}, and are essentially subspaces of $\bigotimes_{i=1}^{n} \M{U}_i$. Note that any subspace of $\bigotimes_{i=1}^{n} \M{U}_i$ of rank $r$ can be written as $(\bigotimes_{i=1}^{n} \M{U}_i )\M{B}$ where $\M{B}$ is a rank $r$ matrix. This matrix is also referred to as the \emph{transfer matrix}. The Hierarchical Tucker subspace construction endows the matrix $\M{B}$ with an additional Kronecker structure, which in turn corresponds to a dimension tree. 

% The difference between a Hierarchical Tucker and a normal Tucker subspace is the core tensor. In hierarchical, the core tensor bears the information of structured hierarchy and is not any random tensor. To clarify we introduce two structures in the next section and describe the format of the core tensors for Hierarchical Tucker.

%\begin{defn}
%Dimesnion Tree(\cite{LarsTree}): A dimension tree $\textit{T}$ with $Root(\textit{T})=\{1,\cdots, n \}$ and depth $\lceil \log_2(d):= min\{i \in \mathbb{N}|i\geq \log_2(d) \}$ that node t is either :\\
%\begin{itemize}
%\item A leaf and singleton $t=\{\mu\}$ on level $l \in \{p-1, p \}$ or
%\item The union of two disjoint successors $S(t)=\{ s_1,s_2\}$ 
%\end{itemize}
%\end{defn}
%\begin{defn}
%The level $l$ of the tree is the set of all nodes that have equal distance to the root:
%\begin{equation}
%	\textit{T}_{I}^l := \{t \in \textit{T}| level(t)=l \}
%\end{equation}
%\end{defn}
%
%The hierarchy in hierarchical subspaces can be shown by a tree as shown in figure(\ref{fig: HT}) and (\ref{fig:TT}).\\
% 
%The two most common trees are as follows:
%
%\begin{itemize}
%\item Balanced tree
%\item Tensor Train
%\end{itemize}
%In what follows we show the relation of subspaces and tensors in different tree structures.
%
%\subsection{Balanced Tree}
%
%This tree takes each higher-subspace to come from two lower-subspaces that are equally weighted in terms of the number of hidden subspaces. W.L.O.G we can say that each level is separated into $\{1,...,\lfloor n/2\rfloor\}$ and $\{\lfloor n/2\rfloor +1, ..., n\}$ where $n$ defines the number of subspaces. \\

\begin{figure}[htbp]
\begin{subfigure}{.49\linewidth}
\centering
	\includegraphics[scale=.49,left]{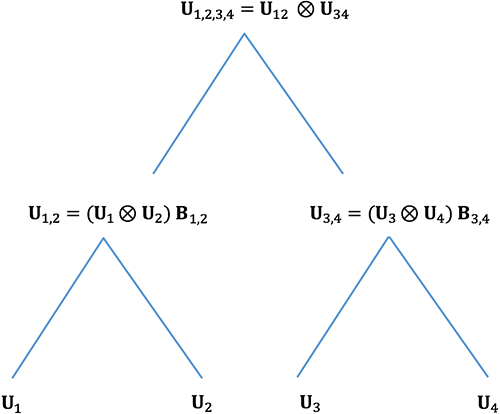}
	\caption{\textcolor{black}{Balanced} tree.}
	\label{fig:selector}
	\end{subfigure}
\begin{subfigure}{.49\linewidth}
\centering
	\includegraphics[scale=0.42, right]{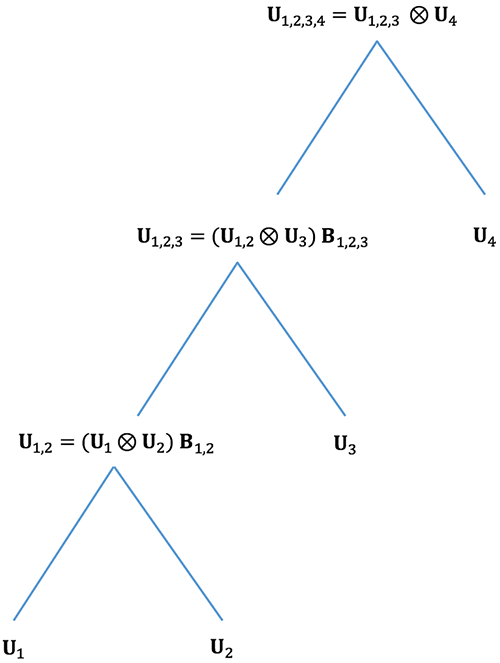}
	\caption{Tensor Train tree.}
	\label{fig:TT}
\end{subfigure}
\caption{Different structure of Trees}
\end{figure}

%For example, a \emph{balanced} Hierarchical Tucker subspace construction based on what is referred to as the balanced dimension tree shown in figure (\ref{fig:selector}) begins with base subspaces specified by $\M{U}_i \in \mathbb{R}^{I_i \times r_i}$ for  $i=1,..,4$. 
For example, if the transfer matrix $\M{B}$ can be further written as $\M{B} = \M{B}_{1,2} \otimes \M{B}_{3,4}$ where $rank(\M{B}_{1,2}) = r_{1,2}$ and $rank(\M{B}_{3,4}) = r_{3,4}$, then,
\begin{align}
\M{u}_{1,2,3,4}= (\M{U}_1\otimes \M{U}_2\otimes \M{U}_3\otimes \M{U}_4)(\M{b}_{1,2} \otimes \M{b}_{3,4})
%\M{u}_{1,2,3,4}=((\M{U}_1\otimes \M{U}_2)\M{b}_{1,2})\otimes ((\M{U}_3\otimes \M{U}_4)\M{b}_{3,4})
\end{align}
By using the Kronecker product property $(\M{a}\M{d}) \otimes (\M{b}\M{c})=(\M{a} \otimes \M{b})(\M{d} \otimes \M{C}) $ one can rewrite the above equation as follows. 
\begin{equation}
\label{eq:HT}
\M{u}_{1,2,3,4}=((\M{U}_1\otimes \M{U}_2)\M{b}_{1,2})\otimes ((\M{U}_3\otimes \M{U}_4)\M{b}_{3,4})
\end{equation}
From this expression it can be seen that the Hierarchical Tucker subspace defined by $col-span(\M{u}_{1,2,3,4})$ corresponds to and is defined with respect to the balanced dimension tree as shown in figure (\ref{fig:selector}).

The same approach could be applied to another dimension tree as shown in figure (\ref{fig:TT}), referred to as the Tensor Train tree \cite{Oseledets_SIAM2011}. In this construction each higher-subspace comes from two lower-subspaces; $\M{U}_{i} $ where $i\in \textsl{T}=\{1,..,n\}$ and the second subspace $\M{U}_{j} $ where $j\in \textsl{T}^\prime=\{1,..,n\}\backslash i$. Backslash means that $i$ and $j$ are mutually exclusive. Corresponding to this tree we note that,
\begin {align*}
	\M{u}_{1,2,3,4} &=\M{U}_{1,2,3}\otimes \M{U}_4 =((\M{U}_{1,2}\otimes \M{U}_{3})(\M{b}_{1,2,3}\otimes \M{I})\otimes \M{U}_4) \\
	&=(((\M{U}_1\otimes \M{U}_2)\M{b}_{1,2}) \otimes \M{U}_3)(\M{b}_{1,2,3}\otimes \M{I})\otimes \M{U}_4\\
	&= (\M{U}_1\otimes \M{U}_2\otimes \M{U}_3\otimes \M{U}_4)(\M{b}_{1,2} \otimes \M{I}) (\M{b}_{1,2,3} \otimes \M{I})
\end {align*}
In general a dimension tree is defined as follows,
\begin{defn}
Dimension Tree\cite{gras-hsvd-2010}: A (binary) dimension tree $\mathfrak{T}$ is a tree whose nodes are represented by a set $S$, that is a set of subsets of $[n]$ with root corresponding to the set $[n]$ and $n$ leaf nodes $\{i\}$ corresponding to the base subspaces $\mb{S}_i, i = 1,2,...,n$. Every node $s \in S$ that is not a leaf has two sons $s_1,s_2$ that form an ordered partition of $S$, i.e. $s_1 \cup s_2 = S, s_1 \cap s_2 = \phi$.
\end{defn}
For example,  the balanced HT tree for an order-4 tensor can be defined as,
\[
S = \{(1, 2, 3, 4), (1, 2), 1, 2, (3, 4), 3, 4\},
\] 
Here the node $(3,4)$ is the parent of $3$ and $4$, $(1,2,3,4)$ is the parent of $(1,2)$ and $(3,4)$ and so on. We will use subscript to denote the transfer matrix corresponding to the nodes in a tree. For example, in the balanced tree example, $\M{B}_{2,3}$ denotes the transfer matrix for node $(2,3)$.

\section{Estimating Hierarchical Subspaces}
\label{sec:3}
%Elements of vectors, matrices or tensors are denoted by adding subscripts to them, for example for a tensor we denote the element by $\T{B}_{i_1i_2 \cdots i_n}.$ 
We are given $N$ data points $\T{x}_i \in \mathbb{R}^{I_1 \times \cdots \times I_n}, i \in [N]$ and a \textcolor{black}{dimension} tree $\mathfrak{T}$ and the ranks corresponding to the nodes in the tree. The problem is to find the best HT subspace, i.e. estimate $\M{U}_j \in \mathbb{R}^{I_j\times r_j}, j \in [n]$ and the \emph{transfer matrices} $\M{B}_s$ for all $s \in S$, that fits the data well in terms of least squares error in projection.

%$ \T{B}  \in \mathbb{R}^{r_1 \times \cdots \times r_n \textcolor{red}{\times N}}$ such that for $\T{x}_i$:
%%$$vec(\T{X}_i)=\sum_{j_n=1}^{I_n} \cdots \sum_{j_1=1}^{I_1} \T{B}_{j_1j_2j_3\cdots j_ni}((\M{U}_1)_{j_1}\otimes \cdots \otimes (\M{U}_n)_{j_n}) $$
%\begin{align}
%&vec(\T{X}_i)=\left(\bigotimes_{i=1}^{n} \M{U}_i\right) vec(\T{B}(:,\cdots,:,i)) \\
%&s.t. \text{  } \T{B}(:,\cdots,:,i) \text{ bears the structure imposed by }\mathfrak{T},
%\end{align}
%where each slice  $\T{B}(:,\cdots,:,i)$ of the tensor $\T{B}$ is obtained by keeping the last index fixed, is constrained to have the Kronecker structure imposed by the dimension tree $\mathfrak{T}$. Here $vec(\cdot)$ denotes the vectorization operation that is the inverse of the \emph{reshaping} operating considered before. See \cite{xx} for more details.

\textcolor{black}{We note that estimating both the dimension tree and the subspaces is a hard problem as there are combinatorially many dimension trees possible.  In this paper we restrict ourselves to the balanced trees, which with slight abuse of notation will be  referred to as the Hierarchical Tucker (HT), and the tree corresponding to Tensor Train (TT). We now present an algorithm for estimating the Hierarchical Subspace. Further finding the best HT subspace approximation is also computationally hard and in this paper we consider a suboptimal algorithm.}

\textbf{\HMSL Algorithm}:
%Before we state the algorithm we need two more definitions.
%\begin{defn}
%\textcolor{black}{Dimension Tree\cite{gras-hsvd-2010}: A dimension tree $\mathfrak{T}$ with $Root(\mathfrak{T})=\{1,\cdots, n \}$ is represented by $S$, a set of subsets of ${\{1,2, \cdots, n}\}$,
%\[
%S = \{(1, 2, 3, 4), (1, 2), 1, 2, (3, 4), 3, 4\},
%\] 
%we name the nodes of a tree via its roots. For example, $(3,4)$ is the parent of $3$ and $4$, or $(1,2,3,4)$ is the parent of $(1,2)$ and $(3,4)$ in a balanced tree. We use subscript to denote the transfer matrix corresponding to the nodes in a tree. For example, $\M{B}_{2,3}$ is the transfer matrix corresponding to the node named $(2,3)$.}\\
%\end{defn}
Before we describe the algorithm we need one more definition.
\begin{defn} \label{def:unfold} (see \cite{gras-hsvd-2010, USCHMAJEW2013133, KoldaMatrix}) \textbf{Unfolding}: Let $s_1$ and $s_2$ be a partition of $[n]$. For an order-$n$ tensor $\T{x} \in \mathbb{R}^{I_1 \times \cdots \times I_n}$, unfolding along $s_1$, denoted by $Unfold(\T{X})=\M{X}^{(s_1)}$ is a matrix of size $I_{s_1} \times I_{s_2}$ where,
	\begin{equation}
	I_{s_1}={\prod_{i \in s_1}I_i} \quad I_{s_2}={\prod_{j \in s_2}I_j},
	\end{equation}
	obtained by lexicographically combining the indices belonging to $s_1$ into row indices, and those belonging to $s_2$ into column indices. 	
%	The mapping indices of a tensor to a matrix and MATLAB codes for this purpose can be found at \cite{KoldaMatrix}. One specific case of unfolding a tensor is vectorizing it to a column vector which we denote by $vec(\T{X})=\V{x}$.
\end{defn}
Algorithm (\ref{alg:HMSL2}) is a simple variant of the Hierarchical SVD \cite{Grasedyck_SIAM2010} computing H-Tucker representation of a single datum. The algorithm takes $N$ tensors $\T{X}_1,\cdots,\T{x}_n \in \mathbb{R}^{I_1 \times \cdots I_n}$, which belong to one class of data as input. Using the specified dimension tree, the algorithm computes the hierarchical subspace, returned as the column span of the matrix $\M{H}$, using a Depth First Traverse (leaves to root) on the dimension tree. The subspaces corresponding to each node are computed using a truncated SVD on the node unfolding and the transfer tensors are computed using the projections on the tensor product of subspace of the node's children (except the root). 
% It also takes a tree and hyper-parameters $r_i$ as the rank of each subspace corresponding to the every node of the tree as input. We choose a rank for each node of the tree, except for the root.}
%The algorithm performs a \textcolor{black}{Depth-First-Traverse (DFT)} (or leaves to root) on the tree. \textcolor{black}{For each node we can unfold the tensor data along the node and then project it onto the Kronecker product of the matrices corresponding to the children's subspaces. After projection, we perform truncated SVD on the projected data and return the result as an estimate of the subspace corresponding to that node.}
%At the end, the algorithm returns the matrix $\M{H}$, whose columns span the HT subspace.
% Method 2 algorithm
\vspace{-5mm}
\begin{algorithm}
\begin{algorithmic}	
\STATE \textcolor{black}{INPUT: $\T{X}_1,\cdots,\T{x}_n \in \mathbb{R}^{I_1 \times \cdots I_n}$, $\mathfrak{T}$ the tree, $r$ the rank of each node of the tree except the root.}
\STATE  $S \gets$ Depth-First-Traverse($\mathfrak{T}$)
\STATE $\text{Stack all } \T{X}_i$ to make $\T{X} \in \mathbb{R}^{I_1 \times I_2 \times \cdots \times I_n \times N}$ 
\FOR {s in  $reverse(S)$ }
\IF {$|s|=1$}
	\STATE 	$\M{X}^{(s)} \gets unfold(\T{X}) \text{ along $s$ (see Definition \ref{def:unfold})} $
	\STATE $\M{X}^{(s)} = \M{U}\M{\Sigma}\M{ V}^T$ \%SVD
	\STATE $\M{U}_{s} \gets \M{U}(:, 1:r_s)$
	\STATE save $\M{U}_{s}$
\ELSIF {$s$ not root}
	\STATE {Split $s$ into its children; $s_1$ and $s_2$}
	\STATE $\M{\tilde{U}} \gets \M{U}_{s_2} \otimes \M{U}_{s_1}$
	\STATE $\M{X}^{(s)} \gets unfold(\T{X}) \text{ along s} $
	\STATE $\M{\tilde{X}} \gets \M{\tilde{U}}\M{\tilde{U}}^T \M{X}^{(s)}$
	\STATE $\M{\tilde{X}} = \M{U}\M{\Sigma}\M{ V}^T$
	\STATE $\M{U}_{s} \gets \M{U}(:, 1:r_s)$ \textcolor{black}{\% Top $r_s$ singular vectors}
	\STATE $\M{B}_s \gets \M{\tilde{U}}^T \M{U}_{s}$
	\STATE store $\M{U}_{s}$	and $\M{B}_s$
\ELSIF {$s$ is root}
	\STATE split root into its children $s_1$ and $s_2$
	\STATE $\M{H} = \M{U}_{s_1} \otimes \M{U}_{s_2} $ 
\ENDIF
\ENDFOR
\STATE Return $\M{H}$
\end{algorithmic}
\caption{Hierarchical Tensor subspace learning}
\label{alg:HMSL2}
\end{algorithm}

\vspace{-1.0em}
\section{Tradeoffs between error, storage and projection cost using tensor subspaces}
\label{sec:4}
We now investigate the storage and projection tradeoffs for various tensor subspace representations on some real data sets. In the following experiments, we use the PIE \cite{PIE} and Weizmann \cite{Weizemann} face data sets. 

%\begin{table}
%	\begin{center}
%		\begin{tabular}{|c|c|c|}
%		
%		\hline
%		Data SetName & Image size & Reshaped tensor\\
%		
%		\hline
%		PIE     & $486 \times 640 $ & $ 18 \times 27 \times  32 \times 20 $\\
%		Weizmann & $512 \times 288$ & $16 \times 32 \times 16 \times 18$\\
%		\hline
%		\end{tabular}
%	\caption{Dimensions of data sets}
%	\label{table:sizes}
%	\end{center}
%\end{table}

\textbf{Preprocessing:} We centered all of the images by subtracting the mean of the samples.  For PIE database each picture is of size 486 x 640 and for Weizmann database each image is of size 512 x 288.  
%We represent each picture as a tensor $\T{x} \in \mathbb{R}^{486 \times 640}$ for \textcolor{red}{the} PIE data set and $\T{x} \in \mathbb{R}^{512 \times 288}$ for \textcolor{red}{the} Weizmann data set (table \ref{table:sizes}). 
We reshaped these images to $\T{x} \in \mathbb{R}^{18 \times 27 \times 32 \times 20}$ (PIE data) and $\T{x} \in \mathbb{R}^{16 \times 32 \times 16 \times 18}$ (Weizmann data). In short, we reshape each 2-D matrix to a 4-D tensor. 
We set aside 50 percent of the samples of each subject for testing and 50 percent for training.

\textbf{Procedure}: For training, we select 4 subjects (classes) out of a collection of 18 subjects randomly for each experiment and estimate the Hierarchical Tensor subspace for each subject. For testing and classifying a data point, we project it on the Hierarchical Subspaces computed for each subject and choose the one with the maximal projection energy.

For \textcolor{black}{Hierarchical Tucker}, one can take two different approaches for computing the projection $\M{X}_{pr}$ from a test data point, say, $\T{X}_{0}$, all of which have different projection and storage costs,
\begin{enumerate}
\item \emph{hier approach 1}: $\M{X}_{pr} = \M{U}_{1,2}^\top \M{X}_{0}^{(1,2)} (\M{U}_{3,4})^\top$. Recall that $\M{X}_{0}^{(1,2)}$ is an unfolding of $\T{X}_0$ - c.f. Definition \ref{def:unfold}.\\%$\M{X}_{pr} = \M{X}_{0}^{(1,2)} \times_{1,2} \M{U}_{1,2}^\top \times_{3,4} \M{U}_{3,4}^\top$.\\
%In this method, $\times_{1,2}$ and $\times_{3,4}$ denote \emph{tensor-matrix multiplication (also called mode-k multiplication)} \cite{tensorProduct}. 
We only have to store $\M{U}_{1,2}$ and $\M{U}_{3,4}$. In a case where $\M{U}_{1,2}, \M{U}_{3,4} \in \mathbb{R}^{n^2 \times r\textprime}$, the cost of projection is $n^4r\textprime+n^2r\textprime^2+r\textprime^2$ and the cost of storage is $2n^2r\textprime$.
\item \emph{hier approach 2}: $\M{X}_{pr}= ((\M{U}_1 \otimes \M{U}_2) \M{B}_{1,2})^\top \M{X}_{0}^{(1,2)}((\M{U}_3 \otimes \M{U}_4) \M{B}_{3,4})^\top$\\
% 
%$\T{X}_{pr}= \T{X}_{0} \times_{1,2}((\M{U}_1 \otimes \M{U}_2) \M{B}_{1,2}) \times_{3,4} ((\M{U}_3 \otimes \M{U}_4) \M{B}_{3,4})$\\
In this method, we store the leaf subspaces and the transfer matrices and then build $\M{U}_{1,2}$ and $\M{U}_{3,4}$. In a case where all of the leaf subspaces are in $\mathbb{R}^{n \times r}$ and the transfer matrices are in $\mathbb{R}^{r^2 \times r\textprime}$, the cost of projection is $n^4r\textprime+n^2r\textprime^2+r\textprime^2+2n^2r^2+2n^2r^2r\textprime$ and the cost of storage is $4nr+2r^2r\textprime$.
\end{enumerate}
% $\T{X}_{pr} = \T{X}_{0} \times_{1,2,3} \M{U}_{1,2,3}^\top \times_{4} \M{U}_{4}^\top$
For the Tensor Train, we use multiplication in the following format  $\M{X}_{pr} = \M{U}_{1,2,3}^\top \M{X}_{0}^{(1,2,3)}\M{U}_{4}^\top$, where we only have to store $\M{U}_{1,2,3}$ and $\M{U}_{4}$. In a case where $\M{U}_{1,2,3} \in \mathbb{R}^{n^3 \times r\textprime}$ and $\M{U}_4 \in \mathbb{R}^{n \times r}$, the cost of projection is  $n^4r+ n^3 r r\textprime$ and the cost of storage is $n^3r\textprime+nr$. For Tucker, we only store the subspaces, For example, in a case where all of the subspaces are in  $\mathbb{R}^{n \times r}$, the cost of projection is $n^4r+n^3r^2+n^2r^3+nr^4+r^4$ and the cost of storage is  $4nr$.

After computing the projection of each test point onto the subspaces corresponding to each class, we label the test tensor with the class that has the largest Frobenius norm of the projection.
We use,
\[
error = \frac{\# \text{misclassified points}}{ \# \text{total test points}}
\]
as the measure of misclassification. We repeat each experiment \textcolor{black}{(selecting the subject randomly from the set of 18 subjects)} 10 times and average over the results.

%%%%%%%%%%%%%%%%%%%%%%%%%%%%%%%%%%%%
% Complexities for PIE
%\begin{table}
%\begin{subtable}{.5\textwidth}
%\begin{tabular}{ |c|c|c|}
%	\hline
%	Algorithm/Cost  & projection & Storage \\ 
%	 \hline \hline
% 	Tucker & $C_{Projection}^{Tucker}$ &  $C_{Storage}^{Tucker}$\\
% 	\hline 
% 	 H-Tucker:App1 & $311040r\textprime+361r\textprime^2$ & 			 $1126r\prime$ \\  
%
%H-Tucker: App2 &  $622080+79315r\textprime+487r\textprime^2$ & $2272+352r\textprime$\\   
% \hline
% TT &   $180+11264r\prime$& $17280 r\prime$\\
% \hline
%	\end{tabular}
%	\caption{The Costs for PIE data set}
%\end{subtable}
% 
%\begin{subtable}{.5\textwidth}
%	\begin{tabular}{ |c|c|c|}
%	\hline
%	Algorithm/Cost & projection & Storage \\ 
% 	\hline \hline
% 	Tucker &  $C_{Projection}^{Tucker}$ &  $C_{Storage}^{Tucker}$\\
% 	\hline 
% 	 H-Tucker:App1 & $180224r\textprime+353r\textprime^2$ & 	$864r\textprime$ \\  
%
%	H-Tucker:App2 &  $261440+441664r\prime+513r\textprime^2$ & 	$1660+620r\textprime$\\    
% \hline
% TT  &$2018500 +  126100 r\textprime$ & $180+11264r\textprime$ \\
% \hline
%\end{tabular}
%\caption{The Costs for PIE data set}
% \end{subtable}
% \caption{The tables demonstrate the costs of storage and testing of each data point with different \textcolor{red}{Hierachical Tucker} and Tucker subspaces for \textcolor{red}{the}  data sets. r is the rank of leaf-level subspaces and $r\textcolor{red}{\textprime}$ denotes the rank of $\M{U}_{1,2}$ and $\M{U}_{3,4}$.}
% \label{table:cost}
%\end{table}
%\vspace{-.4em}

\textbf{Cost Comparison} - 
\begin{figure}
\begin{subfigure}{\linewidth}
\centering
\includegraphics[scale=0.18]{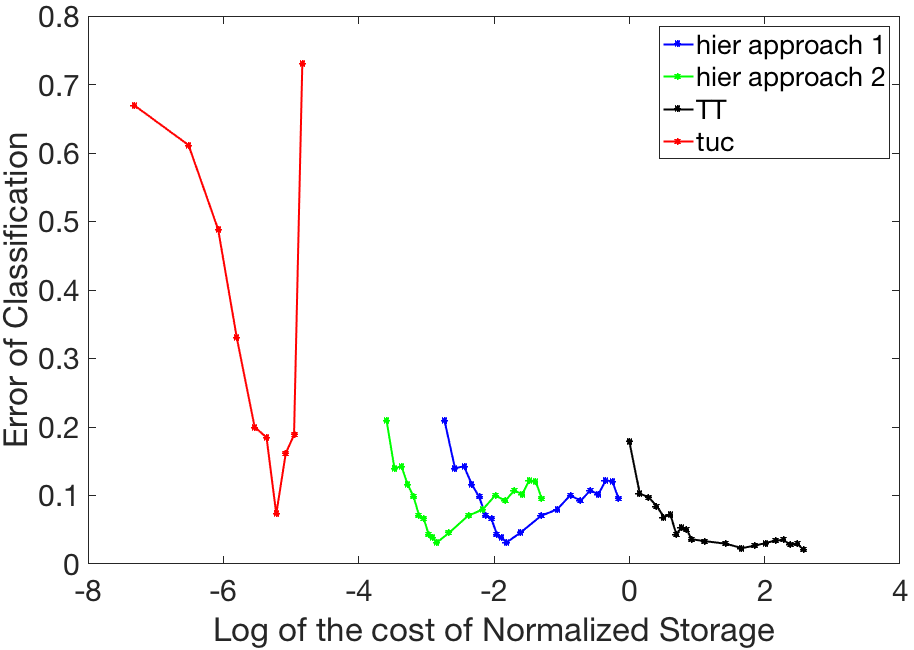}
\caption{Error versus varying cost of storage.}
\label{fig:PIEstorageCost}
\end{subfigure}
\begin{subfigure}{\linewidth}
\centering
\includegraphics[scale=.18]{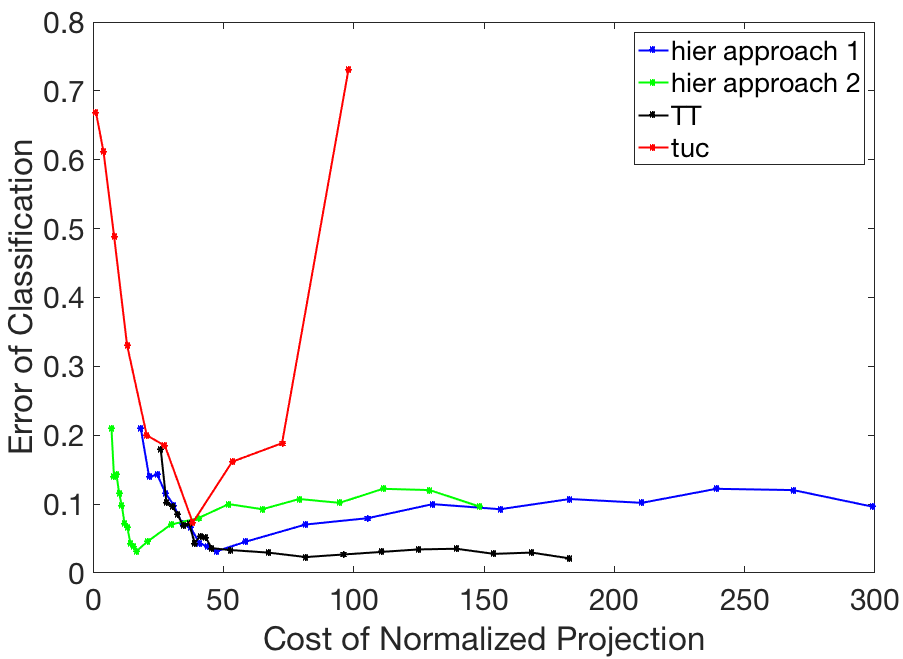}
\caption{Error versus varying cost of projection.}
\label{fig:PIEprojectionCost}
\end{subfigure}
\caption{Results for PIE Data Set}
\end{figure}
% For Weizmann
\begin{figure}
\begin{subfigure}{\linewidth}
\centering
\includegraphics[scale=0.18]{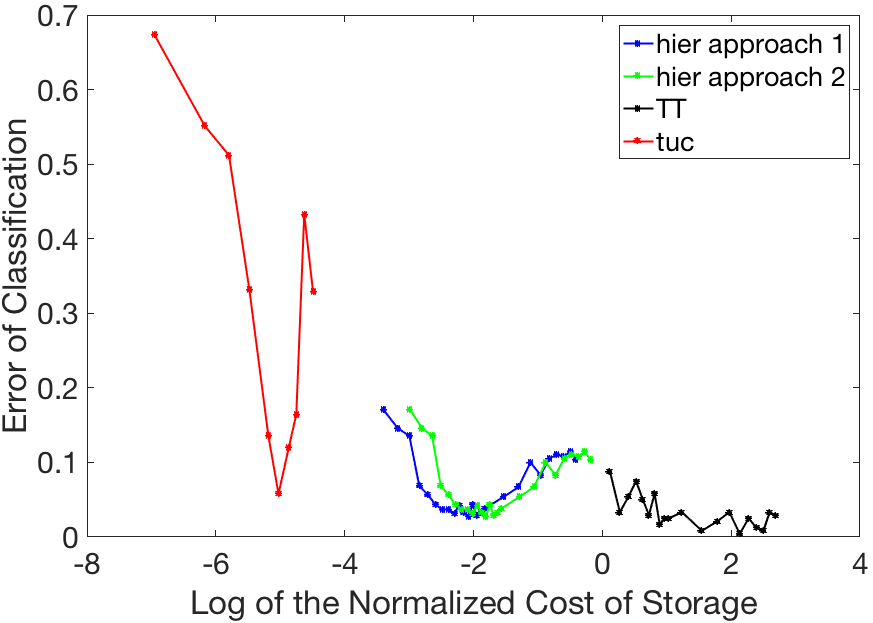}
\caption{Error versus varying cost of storage.}
\label{fig:WeizstorageCost}
\end{subfigure}
\begin{subfigure}{\linewidth}
\centering
\includegraphics[scale=0.18]{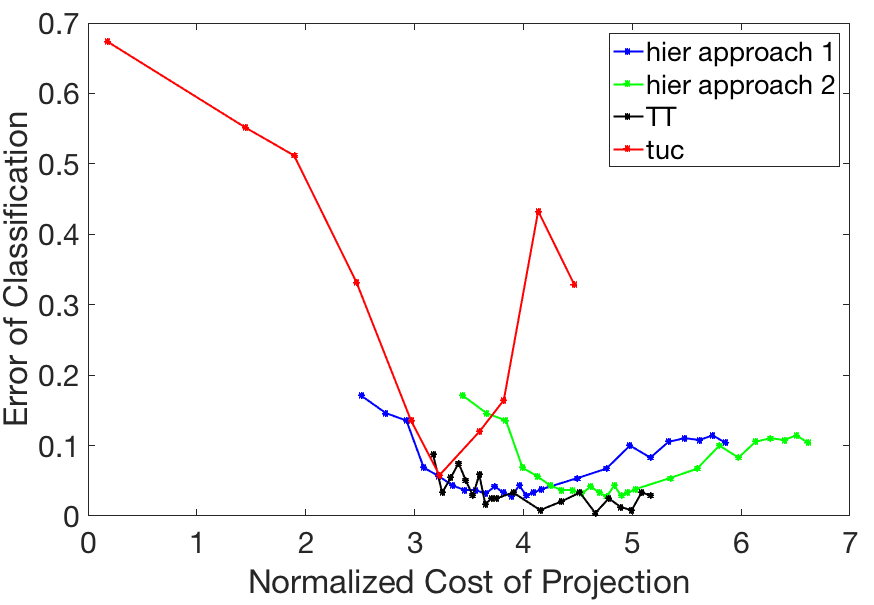}
\caption{Error versus varying cost of projection.}
\label{fig:WeizprojectionCost}
\end{subfigure}
\caption{Results for Weizmann Data Set}
\end{figure}
In the following experiments, we varied $r_i, i \in{1,2,3,4}$ from 10\% to 100\% of the full rank. For Hierarchical and Tensor Train (TT), we set the leaf level subspaces to 70\% of the full rank and varied $r\textprime=r_{1,2}=r_{3,4}$. As seen in the figures (\ref{fig:PIEstorageCost}, \ref{fig:PIEprojectionCost}, \ref{fig:WeizstorageCost}, and \ref{fig:WeizprojectionCost}), we plotted the error rates versus the cost of normalized : 1. Storage 2. Projection. We normalized the costs by dividing them by the number of the dimensions of the vectorized tensors. We can see that different limitations in storage or projection play role in choosing which algorithm performs better. 

In figures (\ref{fig:PIEstorageCost}) and (\ref{fig:WeizstorageCost}), we observe that the cost of storage of the Tucker subspace representation is smaller than Hierarchical Tucker and Tensor Train, however the error rates of Hierarchical Tucker and Tensor Train are much smaller. 

We can also observe that using the Tucker subspace leads to a very strong overfitting for higher ranks. This is due to using large ranks which brings about extra complexity. The bad performance of using the Tucker subspace representation at higher ranks demonstrates that the Tucker method is sensitive to noise, however Hierarchical Tucker is much more robust and Tensor Train demonstrates no overfitting at all. 

In figures (\ref{fig:PIEprojectionCost}) and (\ref{fig:WeizprojectionCost}), we observe that the cost of projection for HT and TT are almost the same as the Tucker representation. We also observe that the classification error of Tensor Train is smaller than the Tucker method at any given computation cost.\\

\begin{figure}
\begin{subfigure}{\linewidth}
\centering
\includegraphics[scale=.19]{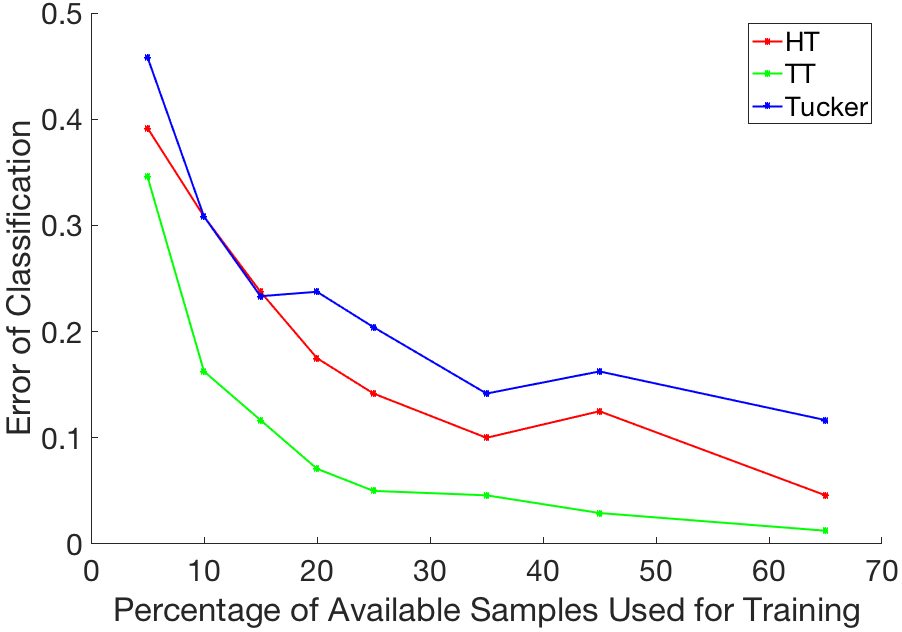}
\caption{Weizmann Data Set Learning Curve}
\label{fig:learningWeiz}
\end{subfigure}
\begin{subfigure}{\linewidth}
\centering
\includegraphics[scale=.19]{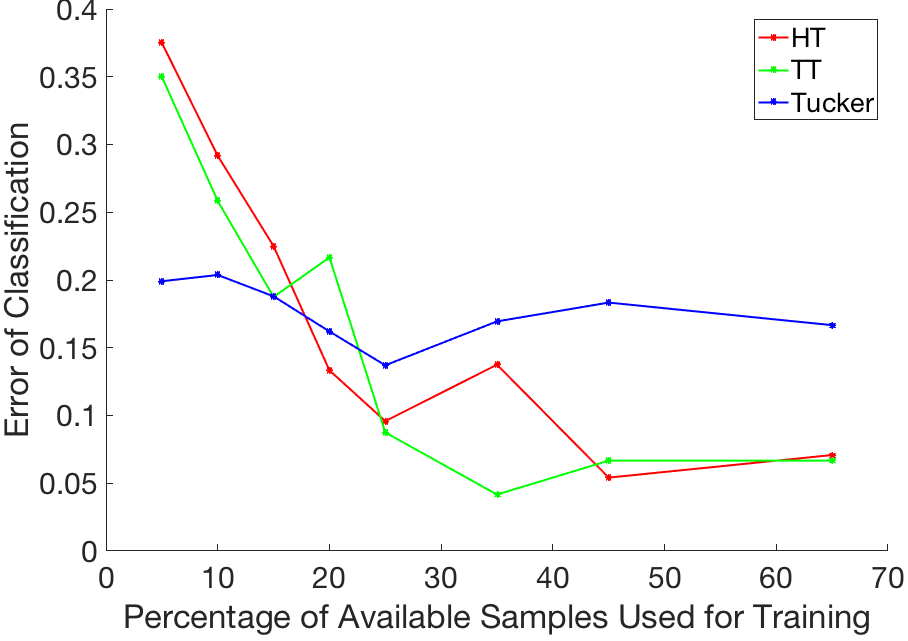}
\caption{PIE Data Set Learning Curve}
\label{fig:learningPIE}
\end{subfigure}
\caption{Results for sample complexity.}
\end{figure}

\textbf{Sample complexity vs error}: We further evaluate the methods for sample complexity, i.e. the number of samples required \textcolor{black}{to achieve} a given classification error. As demonstrated in figures (\ref{fig:learningWeiz}) and (\ref{fig:learningPIE}), Hierarchical methods (TT and HT) tend to perform better in a sense that they need fewer points in order to achieve the same classification error. Among Hierarchical Subspace models, the Tensor Train performs better compared to Hierarchical Tucker (corresp. to the balanced tree). This is particularly interesting, since it shows how the choice of the tree can affect the performance.

\IEEEpeerreviewmaketitle
\bibliographystyle{IEEEbib}
\clearpage
\bibliography{References2}
\end{document}